\documentclass{article} %
\usepackage{iclr2015mod,times}
\usepackage{hyperref}
\usepackage{url}
\usepackage{helvet}
\usepackage{courier}
\usepackage{latexsym}
\usepackage{amsmath,amsthm,amssymb}
\usepackage{graphicx}
\usepackage{multirow}
\usepackage{url}
\usepackage{verbatim}
\usepackage{caption}
\usepackage{subcaption}

\usepackage{color}

\newcommand{\R}{\mathbb{R}}
\newcommand{\shortcite}{\citet}
\renewcommand{\cite}{\citep} 

\title{Incorporating Both Distributional and Relational Semantics in Word Representations}

\author{
Daniel Fried\thanks{Currently at the University of Cambridge.}\\
Department of Computer Science\\
University of Arizona \\
Tucson, Arizona, USA \\
\texttt{dfried@email.arizona.edu} \\
\And
Kevin Duh\\
Graduate School of Information Science \\
Nara Institute of Science and Technology \\
Ikoma, Nara, JAPAN \\
\texttt{kevinduh@is.naist.jp} \\
}

\iclrfinalcopy %

\begin{document}
\maketitle

\begin{abstract}
We investigate the hypothesis that word representations ought to incorporate both distributional and relational semantics. To this end, we employ the Alternating Direction Method of Multipliers (ADMM), which flexibly optimizes a distributional objective on raw text and a relational objective on WordNet. Preliminary results on knowledge base completion, analogy tests, and  parsing show that word representations trained on both objectives can give improvements in some cases.

\end{abstract}

\section{Introduction}

We are interested in algorithms for learning {\em vector representations} of words. 
Recent work has shown that such representations, also known as word embeddings, can successfully capture the semantic and syntactic regularities of words \cite{mikolov13queen} and improve the performance of various Natural Language Processing systems, including information extraction \cite{turian10word,wang13sequence}, parsing \cite{socher13parsing}, and semantic role labeling \cite{collobert11scratch}.

Although many kinds of representation learning algorithms have been proposed so far, 
they are all essentially based on the same premise of {\em distributional semantics} \cite{harris54}, embodied by 
J. R. Firth's dictum: ``You shall know a word by the company it keeps.''
For example, the models of \cite{bengio03neurallm,schwenk07cslm,collobert11scratch,mikolov13distributed,mnih13word} train word representations by exploiting the context window around the word.
Intuitively, these algorithms learn to map words with similar context to nearby points in vector space.  

However, distributional semantics is by no means the only theory of word meaning. {\em Relational semantics}, exemplified by WordNet \cite{miller95wordnet}, defines a word by its relation with other words. Relations such as synonymy, hypernymy, and meronymy \cite{cruse86semantics} create a graph that links words in terms of our world knowledge and psychological predispositions.  For example, stating a relation like ``dog is-a mammal'' gives a precise hierarchy  between the two words, in a way that is very different from the distributional similarities observable from corpora.
Arguably, the vector representation of ``dog'' ought be close to that of ``mammal'', regardless of their distributional contexts. 

We believe {\em both} distributional and relational semantics are valuable for word representations. 
Our goal is to explore how to combine these complementary approaches into a {\em unified} learning algorithm. We thus employ a general representation learning algorithm based on the Alternating Direction Method of Multipliers (ADMM) \cite{boyd2011distributed} for jointly optimizing both distributional and relational objectives. Its advantages include (a) flexibility in incorporating arbitrary objectives, and (b) relative ease of implementation. 

In the following, we first discuss objectives for {\em independently} learning distributional semantics (\S \ref{language-modeling-objective}) or relational semantics (\S \ref{wordnet-modeling-objective}). The ADMM framework that optimizes both objectives is described in \S \ref{joint-objective} and analyzed in \S\ref{experiments}.  To test whether our embeddings are widely applicable, we evaluate three specific ADMM instantiations (each using different ways of incorporating relational semantics) on a wide range of tasks (\S\ref{tasks}).

\section{Objectives for Representation Learning}
\label{model}

\subsection{Distributional Semantics Objective}
\label{language-modeling-objective}

A standard way to implement distributional semantics in representation learning is the Neural Language Model (NLM) of \shortcite{collobert11scratch}. Each word $i$ in the vocabulary is associated with a $d$-dimensional vector $\mathbf{w}_i \in \R^d$, the word's \emph{embedding}. An $n$-length sequence of words $(i_1, i_2, \dots, i_n)$ is represented as a vector $\mathbf{x}$ by concatenating the vector embeddings for each word, $\mathbf{x} = [\mathbf{w}_{i_1};\mathbf{w}_{i_2}\ldots;\mathbf{w}_{i_n}]$. This vector $\mathbf{x}$ is then scored by feeding it through a two-layer neural network with $h$ hidden nodes:
\begin{equation}
    \small
    S_{NLM}(\mathbf{x}) = \mathbf{u}^{\top}(f(\mathbf{Ax} + \mathbf{b}))
\end{equation}
where $\mathbf{A} \in \R^{h \times (nd)}$ is the weight matrix and $\mathbf{b} \in \R^h$ is the bias vector for the hidden layer, $\mathbf{u} \in \R^h$ is the weight vector for the output layer, and $f$ is the sigmoid $f(t) = 1/(1 + e^{-t})$.%

The layer parameters and word embeddings of this model are trained using noise contrastive estimation \cite{smith2005contrastive,gutmann2010noise,mnih13word}.  A sequence of text from the training corpus is corrupted by replacing a word in the sequence with a random word sampled from the vocabulary, providing an implicit negative training example $\mathbf{x_c}$.  
To train the network so that correct sequences receive a higher score than corrupted sequences, the hinge loss function is optimized:
\begin{equation}
    \small
    \label{hinge-loss}
    L_{NLM}(\mathbf{x}, \mathbf{x}_c)=\max(0, 1 - S_{NLM}(\mathbf{x}) + S_{NLM}(\mathbf{x}_c))
\end{equation}
The word embeddings $\mathbf{w}$ and network layer parameters $\mathbf{A}, \mathbf{u}, \mathbf{b}$ are trained with backpropagation, using stochastic gradient descent (SGD) over n-grams in the training corpus. We are concerned with the learned embeddings and disregard the other network parameters after training. 

\subsection{Relational Semantics Objective}
\label{wordnet-modeling-objective}

Methods for learning word representations based on relational semantics have only recently been explored. 
We first present a simple new objective based on WordNet graph distance (\S \ref{graph-distance}), then discuss two recent proposals that directly model relation types (\S \ref{relational-modeling-objective}).
While our objectives focus on relational semantics in WordNet, they are extensible to other kinds of relational data, including knowledge bases like Freebase.

\subsubsection{Graph Distance}
\label{graph-distance}

In this  approach, we aim to train word embeddings such that the distance between word embeddings in the vector space is a function of the distance between corresponding entities in WordNet. The primary entities in WordNet are synonym sets, or \emph{synsets}. Each synset is a group of words representing one lexical concept. WordNet contains a set of relationships between these synsets, forming a directed graph where vertices are synsets and relationships are edges. The primary relationship is formed by the \textsc{Hypernym} (Is-A) relationship. 

By treating these \textsc{Hypernym} relationships as undirected edges between synsets, we approximate semantic relatedness between synsets as the length of the shortest path between two synsets in the graph. We add a common root node at the base of all hypernym trees so that the synset graph is connected, and adopt the similarity function of \citet{leacock1998combining}:
\begin{equation}
    \small
    SynSim(s_i, s_j) = -\log \frac{len(s_i, s_j)}{2 \times \smash{\displaystyle\max_{s \in WordNet}} depth(s)}
\end{equation}
where $len(s_i, s_j)$ is the length of the shortest undirected hypernym path between synsets $s_i$ and $s_j$ in the graph, and $depth$ returns the distance from the root of the hypernym hierarchy to a given synset.

Since there is a many-to-many relationship between words and WordNet synsets, and embeddings for words, not synsets, are desired, we define the similarity between two words to be the maximum similarity between their corresponding synsets, if both words have associated synsets, and undefined otherwise:
\begin{equation}
    \small
WordSim(i, j) = \max_{s_i \in syn(i), s_j \in syn(j)} SynSim(s_i, s_j)
\end{equation}
where $syn(i)$ is the set of synsets corresponding to word $i$.

To integrate WordNet similarity with word embeddings, we define the following Graph Distance loss, $L_{GD}$. For a word pair $(i,j)$, we encourage the cosine similarity between their embeddings $\mathbf{v}_i$ and $\mathbf{v}_j$ to match that of a scaled version of $WordSim(i,j)$:
\begin{equation}
    \label{eqn:loss_gd}
    \footnotesize
    L_{GD}(i, j) = \left(\frac{\mathbf{v}_i \cdot \mathbf{v}_j}{||\mathbf{v}_i||_2||\mathbf{v}_j||_2}  - \left[a\times WordSim(i, j) + b\right]\right)^2
\end{equation}
where $a$ and $b$ are parameters that scale $WordSim(i,j)$ to be of the same range as the cosine similarity between embeddings.

SGD is used to train the word embeddings as well as the scalar parameters $a$ and $b$. Pairs of words with defined $WordSim$ (i.e. if both words have synsets) are sampled from the vocabulary and used as a single training instance. Details of the sampling process are presented in \S\ref{experiments}.

\subsubsection{Existing Relational Objectives}
\label{relational-modeling-objective}
We are aware of two recent approaches from the Knowledge Base literature which, in addition to representing words (entities) with vector embeddings, directly represent a knowledge base's relations as operations in the vector embedding space.  These models both take as input a tuple $(v_l, R, v_r)$ representing a possible relationship of type $R$ between words $v_l$ and $v_r$, and assign a score to the relationship. 

The TransE model of \citet{bordes2013translating} represents relationships as translations in the vector embedding space. For two words $v_l$ and $v_r$, if the relationship R holds, i.e. $(v_l, R, v_r)$ is true, then the corresponding embeddings $\mathbf{v}_l, \mathbf{v}_r \in \R^d$ should be close after translation by the relation vector $\mathbf{R} \in \R^d$. The score of a relationship tuple is the similarity between $\mathbf{v}_l+\mathbf{R}$ and $\mathbf{v}_r$, measured by the negative of the residual:
\begin{equation}
    \small
    S_{TransE}(v_l, R, v_r) = -||\mathbf{v}_l + \mathbf{R} - \mathbf{v}_r||_2
\end{equation}

\citet{socher2013reasoning} introduce a Neural Tensor Network (NTN) model that allows modeling of the interaction between embeddings using tensors. The NTN model is a two-layer neural network with $h$ hidden units and a bilinear tensor layer directly relating  embeddings. This provides a more expressive model than TransE, but also requires training a larger number of parameters for each relation.
The scoring function for a relation R is
\begin{equation}
    \small
    S_{NTN}(v_l, R, v_r) = \mathbf{U}^{\top}f\left(\mathbf{v}_l^{\top}\mathbf{W}_R\mathbf{v}_{r} + \mathbf{V}_{R} \begin{bmatrix}\mathbf{v}_l \\ \mathbf{v}_r\end{bmatrix} + \mathbf{b}_R\right)
\end{equation}
where $f$ is the sigmoid non-linearity applied elementwise, $\mathbf{U} \in \R^{h}$ is the weight vector of the output layer, and $\mathbf{W}_R \in \R^{d\times d \times h}$, $\mathbf{V}_R \in \R^{h \times 2d}$ and $\mathbf{b}_R \in \R^{k}$ are a tensor, matrix, and bias vector respectively for relationship $R$.

As in the Neural Language Model, embeddings and parameters for these relational models are trained using contrastive estimation and SGD, using the hinge loss as defined in (\ref{hinge-loss}), where $S_{NLM}$ is replaced by either the $S_{TransE}$ or $S_{NTN}$ scoring function on tuples.\footnote{As in the graph distance objective, we must map from synsets to words. In each SGD iteration, a relationship tuple $(s_l, R, s_r)$ is sampled from WordNet such that synsets $s_l$ and $s_r$ contain words in the vocabulary. One word is sampled for each synset from the set of words in the vocabulary contained in the synset, producing a tuple $(w_l, R, w_r)$. This is the correct tuple to be used in training, treating words as entities for the relational model. To produce the corrupted tuple, one of $w_l$, $R$, or $w_r$ is randomly replaced.}

\section{Joint Objective Optimization by ADMM}
\label{joint-objective}

We aim to train a set of word embeddings that, along with the corresponding model parameters, satisfy both the distributional modeling objective (Sec.~\ref{language-modeling-objective}) and one of the relational modeling objectives (Sec.~\ref{wordnet-modeling-objective}). 
We adopt the Alternating Direction Method of Multipliers (ADMM) approach~\cite{boyd2011distributed}. Rather than use the same set of embeddings to evaluate the loss functions for both the distributional and relational objectives, the embeddings are split into two sets, one for each objective, and allowed to vary independently. 
An augmented Lagrangian penalty term is added to constrain the corresponding embeddings for each word to have minimal difference. 
The advantage of this approach is that existing methods for optimizing each objective independently can be re-used, leading to a flexible and easy-to-implement algorithm. 

We describe the ADMM formulation using graph distance as the WordNet modeling objective, but a similar formulation holds when using other relational objectives. Let $\mathbf{w}$ be the set of word embeddings $\{\mathbf{w}_1, \mathbf{w}_2, \ldots \mathbf{w}_{N'}\}$ for the distributional modeling objective, and $\mathbf{v}$ be the set of word embeddings $\{\mathbf{v}_1, \mathbf{v}_2, \ldots \mathbf{v}_{N''}\}$ for the relational modeling objective, where $N'$ is the number of words in the model vocabulary of the corpus, and $N''$ is the number of words in the model vocabularies of WordNet. Let $I$ be the set of $N$ words that occur in both the corpus and WordNet, i.e. the intersection. 
Then we define a set of vectors $\mathbf{y}$ = $\{\mathbf{y}_1, \mathbf{y}_2, \ldots \mathbf{y}_N\}$, which correspond to Lagrange multipliers, to penalize the difference $(\mathbf{w}_i - \mathbf{v}_i)$ between sets of embeddings for each word $i$ in the joint vocabulary $I$:
\begin{equation}
    \label{eqn:augmented_lagrangian}
    \footnotesize
    L_{P}(\mathbf{w}, \mathbf{v}) = \sum_{i \in I} \left (\mathbf{y}_i^{\top} (\mathbf{w}_i - \mathbf{v}_i) \right ) + \frac{\rho}{2} \left ( \sum_{i \in I} ||\mathbf{w}_i - \mathbf{v}_i||_2^2 \right)
    \end{equation}
In the first term, $\mathbf{y}$ has  same dimensionality as $\mathbf{w}$ and $\mathbf{v}$, so a scalar penalty is maintained for each entry in every embedding vector. The second residual penalty term with hyperparameter $\rho$ is added to avoid saddle points. Later we shall see that $\rho$ can be viewed as a step-size during the update of $\mathbf{y}$. 

Finally, this augmented Lagrangian term (Eq. \ref{eqn:augmented_lagrangian}) is added to the sum of the loss terms for each objective (Eq. \ref{hinge-loss} and Eq. \ref{eqn:loss_gd}). Let $\theta = (\mathbf{u}, \mathbf{A}, \mathbf{b})$ be the parameters of the language modeling objective, and $\phi = (a, b)$ be the parameters of the WordNet graph distance objective. The final loss function we optimize becomes:
\begin{equation}
    \label{eqn:joint_loss}
    \small
    L = L_{NLM}(\mathbf{w}, \theta) + L_{GD}(\mathbf{v}, \phi) + L_{P}(\mathbf{w}, \mathbf{v})
\end{equation}

The ADMM algorithm proceeds by repeating the following three steps until convergence: 
\begin{enumerate}
    \item Perform stochastic gradient descent on $\mathbf{w}$ and $\theta$ to minimize $L_{NLM}+L_{P}$, with all other parameters fixed.
    \item Perform stochastic gradient descent on $\mathbf{v}$ and $\phi$ to minimize $L_{GD}+L_{P}$, with all other parameters fixed.
    \item For all embeddings $i$ corresponding to words in both the n-gram and relational training sets, update the constraint vector $\mathbf{y}_i$: 
    \begin{equation}
        \small
    \mathbf{y}_i := \mathbf{y}_i + \rho(\mathbf{w}_i - \mathbf{v}_i)
    \end{equation}
\end{enumerate}
Since $L_{NLM}$ and $L_{GD}$ share no parameters, the gradient descent step for $\mathbf{w}$ in Step 1 does not depend on $\phi$ (the parameters of the relational objective). The derivative of $L_{NLM}(\mathbf{w}, \theta) + L_{P}(\mathbf{w}, \mathbf{v})$ with respect to $\mathbf{w}_i$ is simply the derivative of $L_{NLM}(\mathbf{w}, \theta)$ plus $\mathbf{y}_i + \rho \left (\mathbf{w}_i - \mathbf{v}_i \right)$; the second term acts like a bias term to make  $\mathbf{w}_i$ closer to  $\mathbf{v}_i$.
Similarly, the gradient descent step for $\mathbf{v}$ does not depend on $\theta$, so Step 2 can be optimized easily. In Step 3, a large difference $(\mathbf{w}_i - \mathbf{v}_i)$ causes $\mathbf{y}_i$ to become large, and therefore increases the constraint for the two sets of embeddings to be similar in both Steps 1 and 2.

We note that it is possible to introduce a weight parameter $\alpha \in [0,1]$ into the joint loss function (Eq. \ref{eqn:joint_loss}) to prioritize either $L_{NLM}$ or $L_{GD}$:
\begin{equation*}
        L = \alpha L_{NLM}(\mathbf{w}, \theta) + (1 - \alpha)L_{GD}(\mathbf{v}, \phi) + L_P(\mathbf{w}, \mathbf{v})
\end{equation*}
Empirically we found that while the difference between using joint objective compared to single objective is large, the exact value of $\alpha$ does not significantly change the results; all experiments here use equal weighting.

\section{Data Setup and Analysis}
\label{experiments}
\begin{figure}[ht]
        \centering
          \begin{subfigure}{\linewidth}
        \includegraphics[width=\linewidth]{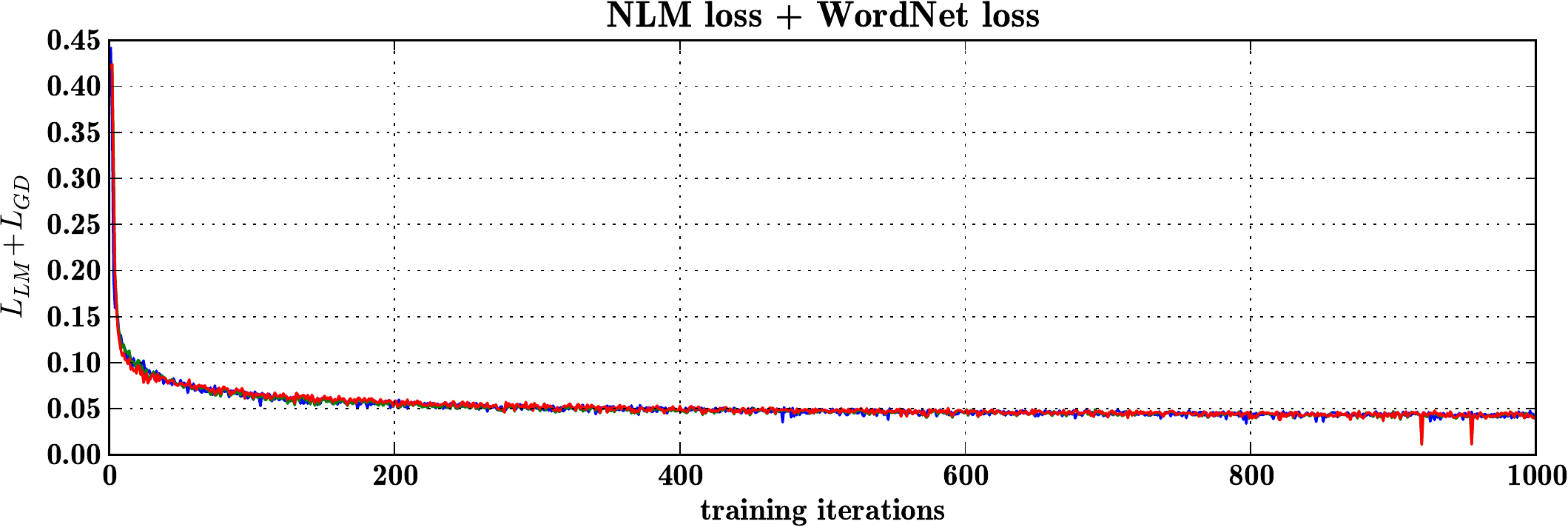}
        \caption{\label{fig:joint_loss}Mean of joint loss without penalty term, $L_{NLM} + L_{GD}$}
        \end{subfigure}
    \begin{subfigure}{\linewidth}
        \includegraphics[width=\linewidth]{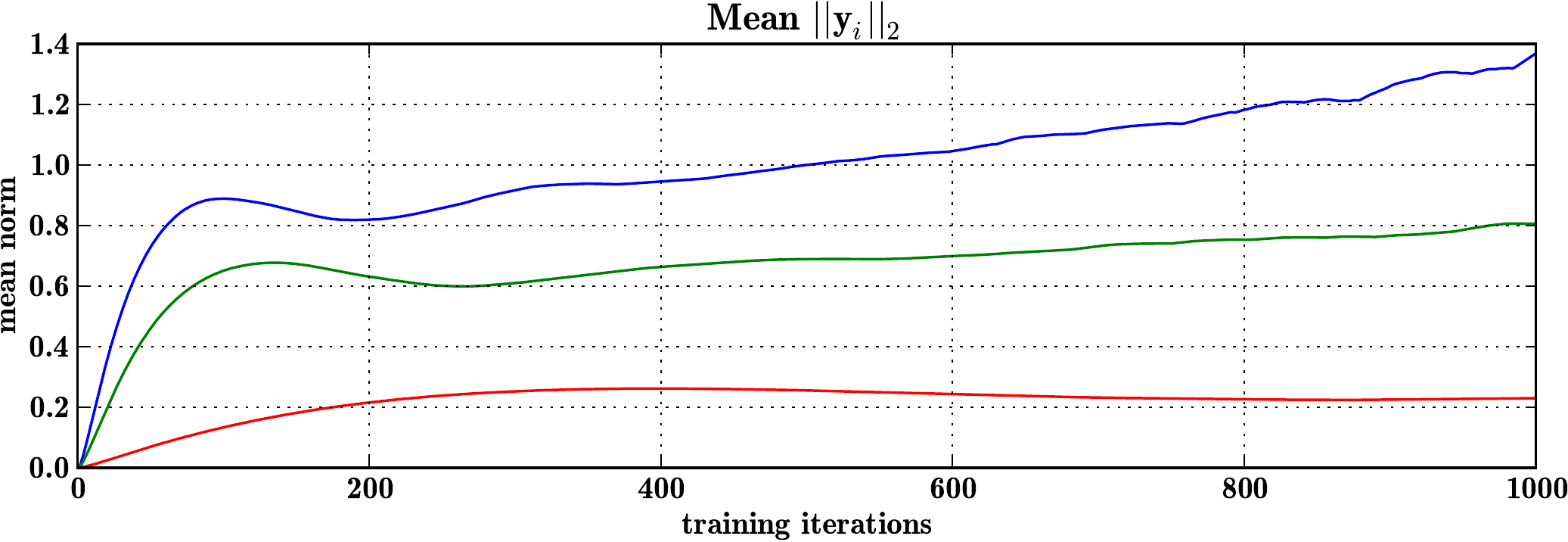}
        \caption{\label{fig:constraint_magnitude}Average magnitude of the constraint vectors, $\mathbf{y}$.}
    \end{subfigure}\\
    \begin{subfigure}{\linewidth}
        \includegraphics[width=\linewidth]{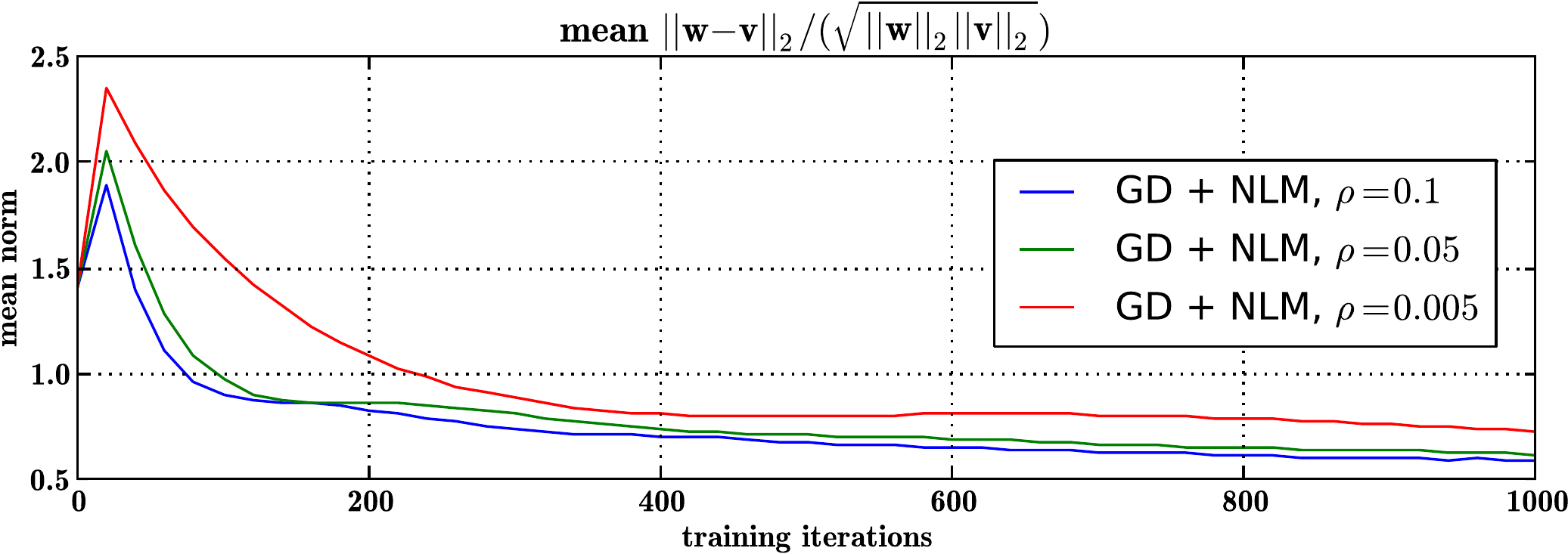}
        \caption{\label{fig:scaled_residual_magnitude}Magnitude of the embedding residuals, scaled by the embedding magnitudes, averaged across all embeddings.}
    \end{subfigure}\\
    \caption{Analysis of ADMM behavior as training iteration progresses, for varying values of $\rho$. 
    }
    \label{fig:admm_stats}
\end{figure}

The distributional objective $L_{NLM}$ is trained using 5-grams from the Google Books English corpus, distributed by UC Berkeley in the Web 1T format\footnote{\url{http://tomato.banatao.berkeley.edu:8080/berkeleylm_binaries/}}. This corpus contains over 180 million 5-gram types and 31 billion 5-gram tokens. In our experiments, 5-grams are preprocessed by lowercasing all words. The top 50,000 unigrams by frequency are used as the vocabulary. All less-frequently occurring words are replaced with a token, \textsc{Rare}, which has its own vector space embedding. In each ADMM iteration, a block of 100,000 n-grams is sampled from the corpus. Each n-gram in the block (and a corrupted, noise-contrastive version of the n-gram, see \S\ref{language-modeling-objective}) is used to perform gradient descent on the distributional loss function $L_{NLM}$ or its ADMM equivalent.

Training data for the relational objective varies depending on whether the graph distance objective (GD) or one of the two relational objectives (NTN or TransE) is used. When the graph distance objective is used, word pairs with defined graph distance (e.g.\ words that are contained in WordNet synsets) are randomly sampled and used as training instances. In each ADMM iteration, 100,000 words are sampled with replacement from the vocabulary. For each sampled word $w$, up to 5 other words $v$ with defined graph distance to word $w$ are sampled from the vocabulary. Each pair $w, v$ is then used as a training instance for gradient descent on the $L_{GD}$ loss function (\S \ref{graph-distance}) or its ADMM equivalent.

When either NTN or TransE is used as the relational objective, WordNet relationship tuples $(s_1, R, s_2)$ where synsets $s_1$ and $s_2$ contain words in the vocabulary, are used as the training instances for stochastic gradient descent using noise contrastive estimation (\S \ref{relational-modeling-objective}). For comparison with the existing work of~\shortcite{socher2013reasoning}, we use their dataset, which contains training, development, and testing splits for 11 WordNet relationships (Table \ref{wordnet_rels}). The entire training set is presented to the network in randomized order, one instance at a time, during each iteration of training.

We next provide an analysis of the behavior of the ADMM joint model ($L_{NLM}+L_{GD}$) on the training set in Figure~\ref{fig:admm_stats}.
Figure~\ref{fig:admm_stats}(a) plots the learning curve by training iteration using varying values of the $\rho$ hyperparameter.
Although establishing convergence guarantees for non-convex loss functions for ADMM is theoretically still an open question (e.g.\ $L_{NLM}$ is a non-convex multi-layer neural net), we empirically observe convergence on our dataset for various values of $\rho$.
Further, in accordance with previous works \cite{boyd2011distributed}, ADMM attains a reasonable objective value relatively quickly in a few iterations; our loss converges around 100 iterations.\footnote{On a 3.3Hz Xeon CPU, this took about 9 hours for ADMM, not more than the 7 hours for independent $L_{NLM}$ and 3 hours for independent $L_{GD}$ objectives combined.} 
Figure~\ref{fig:admm_stats}(b) shows the change in the mean norms of the Lagrange multipliers $||\mathbf{y}_i||_2$. The magnitude of these norms indicates the degree to which the $\mathbf{w}_i$ and $\mathbf{v}_i$ vectors are being constrained in the current ADMM iteration. The norm gradually increases, indicating the tightening of constraints in each iteration. As expected, larger values of $\rho$ lead to faster increases of $||\mathbf{y}_i||_2$. 
Finally, Figure~\ref{fig:admm_stats}(c) shows the normalized difference between the resulting sets of embeddings $\mathbf{w}$ and $\mathbf{v}$, which decreases as desired.\footnote{The reason for the peak around iteration 50 in Figure~\ref{fig:admm_stats}(c) is that the embeddings begin with similar random initializations, so initially differences are small; as ADMM starts to see more data, $\mathbf{w}$ and $\mathbf{v}$ diverge, but converge eventually as $\mathbf{y}$ become large.} 

As we perform SGD on ever-more data, the norms of $\mathbf{w}$ and $\mathbf{v}$ generally increase. A conventional solution is to add the L2 norms of $\mathbf{w}$ and $\mathbf{v}$ as additional regularizers in the objective function. We found that, on the knowledge base completion task (\S\ref{tasks}), L2 regularization decreased the performance of all ADMM models, but slightly increased the performance of the NTN and TransE single objective models; on the analogy test tasks, it decreased the performance of the GD, NLM, and all ADMM models, but increased the performance of NTN and TransE. Since regularization hurt performance for the majority of models, and the evaluation results converge regardless, we use unregularized models in all experiments reported here.

\section{Task-specific Evaluation}
\label{tasks}

We now compare the embeddings learned with different objectives on three different tasks.
For all experiments, we use 50-dimensional embeddings taken from iteration 1000 of training, with $\rho=0.05$ for ADMM.

\subsection{Knowledge Base Completion}
\label{task-kb}

\begin{table*}[t]
    \centering
\small
\begin{tabular}{|l|r|r|r||c|c||c|c|}
\hline
& \multicolumn{3}{c||}{Number of Relationships} & \multicolumn{4}{c|}{Classification Accuracies by Model on Test Set}\\
\hline
Relationship Type &  Train&  Dev&  Test &       NTN &  NTN + NLM &    TransE &  TransE + NLM \\
\hline
{\sc HasInstance}             & 36178 &    1632 &     6334 &  {\bf 76.66} &         76.33 &  79.98 &            {\bf 80.20} \\
{\sc TypeOf}                  & 30556 &    1334 &     5504 &  81.59 &         {\bf 83.79} &  {\bf 85.35} &            84.77 \\
{\sc MemberHolonym}           & 9146  &     614 &     2346 &  88.44 &         {\bf 88.61} &  {\bf 90.57} &            90.36 \\
{\sc MemberMeronym}           & 9223  &     522 &     2268 &  {\bf 85.27} &         83.95 &  81.34 &            {\bf 81.74} \\
{\sc PartOf}                  & 6600  &     334 &     1266 &  {\bf 80.72} &         80.17 &  82.06 &            {\bf 83.80} \\
{\sc HasPart}                 & 6139  &     296 &     1348 &  74.40 &         {\bf 76.33} &  76.48 &            {\bf 77.52} \\
{\sc DomainRegion}            & 4227  &     168 &      592 &  {\bf 68.58} &         68.41 &  70.94 &            {\bf 72.46} \\
{\sc SynsetDomainTopic}     & 3976  &     144 &      622 &  {\bf 91.15} &         90.67 &  89.06 &            {\bf 90.35} \\
{\sc SubordinateInstanceOf} & 3778  &     146 &      650 &  {\bf 94.00} &         92.15 &  91.53 &            {\bf 92.46} \\
{\sc SimilarTo}               & 1659  &       4 &       42 &  {\bf 64.28} &         57.14 &  {\bf 54.76} &            {\bf 54.76} \\
{\sc DomainTopic}             & 1099  &      24 &      116 &  {\bf 67.24} &         62.06 &  68.10 &            {\bf 72.41} \\
\hline
Overall & 112581 & 5218 & 21088 & 80.95 & {\bf 81.27} & 82.87 & {\bf 83.10} \\
\hline
\end{tabular}
\caption{\label{wordnet_rels} Knowledge Base Completion results: counts by WordNet relation type and test classification accuracies for  NTN, TransE, and joint objectives.}
\end{table*}

Models trained using either the NTN or TransE relational modeling objective (\S \ref{relational-modeling-objective}) learn a vector-space representation of relationships in WordNet. We use the methodology and datasets of \citet{socher2013reasoning} to evaluate the models' ability to classify relationship triplets as correct or incorrect. This relation classification is useful for ``completing'' or ``extending'' a knowledge base with new facts. The testing set consists of correct relationship tuples that are present in WordNet, and incorrect tuples created by randomly switching entities from correct tuples. A development set is used to determine threshold scores $T_R$ for each relation that maximize classification accuracy when tuples $(v_l, R, v_r)$ having $S(v_l, R, v_r) \ge T_R$ are classified as correct, and tuples having a score lower than $T_R$ are classified as incorrect.

Models trained using a joint objective (NLM with either TransE or NTN)\footnote{GD is not evaluated since it does not model relation types.} are evaluated by taking the $\mathbf{v}$ set of vectors (those learned for the relational objective) and passing these through the relational objective function to score a given tuple. 
The test accuracies are shown in Table \ref{wordnet_rels}. Overall, the NTN baseline achieves 80.95\% accuracy\footnote{Our NTN results differ from those reported in \cite{socher2013reasoning}, likely due to differences in the optimizer (SGD vs L-BFGS), embedding size, and the use or lack of regularizer.}, while the joint objective NTN+NLM improves it to 81.27\%. Similarly, TransE+NLM (83.10\%) outperforms the TransE baseline (82.87\%).
 We conclude that ADMM can give small albeit noticeable improvement to TransE and NTN. 
 Table \ref{wordnet_rels} also shows the accuracies by relation type. We note that TransE+NLM performs at least as well as the TransE single objective across all categories except {\sc TypeOf} and {\sc MemberHolonym}.

We also experimented with using the average of $\mathbf{w}$ and $\mathbf{v}$ vectors (rather than $\mathbf{v}$ itself after ADMM) for relation classification. This produced lower overall accuracies for the joint model (75.38\% for NTN+NLM  and 71.73\% for TransE+NLM), implying that differences between $\mathbf{w}$ and $\mathbf{v}$ may still be important in actual tasks. 

\begin{figure}[hp]
    \centering
    \begin{subfigure}{\linewidth}
        \includegraphics[width=0.85\linewidth]{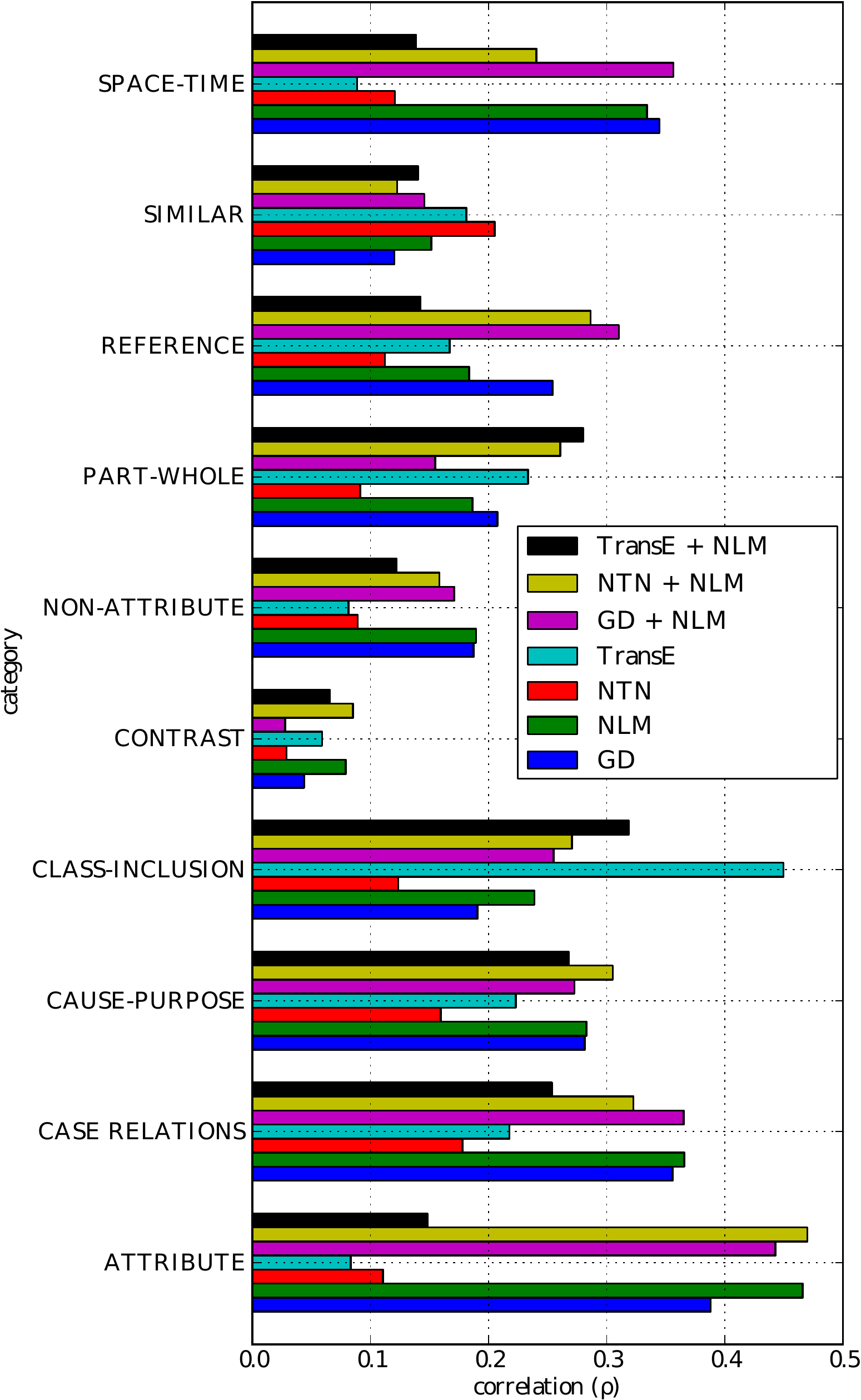}
    \end{subfigure}
    \caption{\label{fig:semeval_bycategory} Detailed comparison of correlation and accuracy for various embeddings on SemEval-2012 Task 2, by relationship category.}
\end{figure}

\subsection{Analogy Tests for Relational Similarity}
\label{task-semeval}

SemEval-2012 Task 2~\cite{jurgens2012semeval} is a relational similarity task similar to SAT-style analogy questions. The task is to score word pairs by the degree to which they belong in a relation class defined by a set of example word pairs. For example, the relation class \textsc{Reverse} contains the example pairs (\texttt{attack,defend}) and (\texttt{buy,sell}). There are 69 testing relation categories, each with three or four example word pairs. In the evaluation, the model is shown a number of testing relation pairs in each category and scores each testing pair according to its similarity to the example relation pairs. These similarity scores are then compared to human similarity judgements. This is an useful task to test whether the positioning of learned embeddings in vector space leads to some meaningful semantics. 
\begin{table}[t]
    \centering
    \begin{tabular}{|l||c|c|}
\hline
Embedding & Accuracy & Correlation \\\hline\hline
NLM &              {\bf 0.42} &                 0.25 \\
\hline
GD &              0.41 &                 {\bf 0.28} \\
GD + NLM  &              0.41 &                 0.25 \\
\hline
NTN &              0.36 &                 0.12 \\
NTN + NLM  &              0.41 &                 0.25 \\
\hline
TransE &              0.37 &                 0.16 \\
TransE + NLM  &              0.38 &                 0.18 \\
\hline
\end{tabular}
\caption{\label{tbl:semeval_scores} Analogy Test results: Comparison of single and joint objective embeddings. For a random baseline, accuracy=.31 and correlation=.018.~\cite{jurgens2012semeval}.}
\end{table}

Following \citet{zhila2013combining}, we evaluate the embeddings in this task on their ability to represent relations as translations in the vector space. A given relation pair $(word_1,word_2)$ is represented as the vector difference between the two words, $\mathbf{w}_2 - \mathbf{w}_1$, where $\mathbf{w}_1$ and $\mathbf{w}_2$ are the embeddings of words $word_1$ and $word_2$. Similarities between the example relations and the relations to be scored are computed using cosine distance of these resulting embedding representations. 
One evaluation metric is the Spearman's correlation coefficient between the similarity scores output by the model and the scores assigned to pairs by human judges. The second metric is the MaxDiff accuracy, which involves choosing both the most similar and least similar example pairs to a given target pair from a set of four or five pre-defined choices. %

A summary of the results is shown in Table \ref{tbl:semeval_scores}. We observe: 
\begin{enumerate}
\item NLM achieves scores competitive with the recurrent neural language models used in \cite{zhila2013combining}, which had a maximum accuracy of 0.41 and correlation of 0.23. 
\item GD by itself also achieves comparable scores, with 0.42 accuracy and 0.28 correlation. It is interesting to note that independent distributional and relational objectives achieve similar results for this task. 
\item The joint objective does not appear to help in this task, however. E.g.,~GD+NLM does not outperform GD; while NTN+NLM outperforms NTN, it does not outperform NLM. 
\end{enumerate}

To analyze this result further, we show the scores by relation category in Figure~\ref{fig:semeval_bycategory}. Despite NLM, GD, GD+NLM, and NTN+NLM achieving similar top scores overall, we observe that the scores by category are considerably varied. 
We conclude our current objectives, either distribution or relational, are too coarse-grained to reliably address the analogy test. 
In particular, the relation categories for this task do not correspond to the relation types on WordNet. Since it is infeasible to expect a large WordNet-like resource that is annotated in the particular categories for this task, an objective that includes some form of unsupervised relation clustering may be necessary. 

\subsection{Dependency Parsing}

Dependency parsing experiments are performed on the SANCL2012 ``Parsing the Web'' data \cite{petrov12sancl}. 
The setup is to train a parser on news domain (Wall Street Journal) and evaluate on out-of-domain web text. Our goal is to see whether the performance of a standard parser can be improved by simply using embeddings as additional features. This can be seen as a kind of semi-supervised feature learning \cite{koo08simple}.

Following \cite{wu13dependency}, we first cluster the embeddings using k-means, then incorporate the cluster ids as features. The reasoning is that discrete cluster ids are easier to incorporate into existing parsers as conjunctions of features. We use the standard first-order MST parser\footnote{\url{sourceforge.net/projects/mstparser/}}. For simplicity, we only attempt to cluster the embeddings into $k=64$ clusters and report results on the development set; a more extensive experiment involving multiple $k$ and model selection is left as future work. %

\begin{table}[th]
\begin{center}
    \setlength\tabcolsep{3pt}%
\begin{tabular}{|l||c||c|c|c|c|c|}
\hline
  Embedding     & AVERAGE          & Answers		& Emails		& Newsgroups 	& Reviews 	& Weblogs \\\hline\hline
None  		& 75.83		& 73.40	  	& 73.72		& 74.88		& 75.46		& 81.70 \\
\hline
NTN			& 75.85		& 73.56		& 73.48		& 74.88		& 75.53		& 81.58 \\
TransE		& 75.86		& 73.30		& 73.69		& 75.09		& 75.68		& 81.74 \\
GD			& 75.90		& 73.54		& 73.73		& 75.03		& 75.65		& 81.55 \\
\hline
NLM			& 76.03       	& 73.65		& 73.77		& 74.96		& 75.81		& 81.94 \\
NLM + NTN 	& 76.14       	& {\bf 73.81}	& 73.87		& 75.09		& {\bf 75.92}	& 82.01 \\
NLM + TransE  	& 76.01       	& 73.50	  	& {\bf 73.89}	& 75.09		& 75.53		& 82.02 \\
NLM + GD 	& {\bf 76.18}	& 73.78		& 73.69		&{\bf 75.39}	& 75.76		& {\bf 82.28} \\  
\hline        
\end{tabular}
\caption{\label{tab:parsing} Comparison of parsers using different embedding features. Labeled Arc Score (LAS) on five web domains and their average are reported. }
\end{center}
\end{table}

Table \ref{tab:parsing} shows the labeled attachment scores (LAS), i.e. the accuracy of predicting both correct syntactic attachment and relation label for each word. We observe: 
\begin{enumerate} 
\item Incorporating embeddings from joint objective training always helps; all of these embeddings improve upon the case of no embeddings (None) in all five domains. This is a nice result considering that our embeddings are trained on Google Books, not SANCL, and have a 9-13\% token out-of-vocabulary rate on the data. 
\item In contrast, improvements from embeddings trained from a single relational objective are mixed, and are in general poorer than those trained from a single distributional objective (NLM). This suggests distributional information may be more effective for this task. 
\item The best results are achieved by the joint models NLM+GD (average LAS of 76.18) and NLM+NTN (76.14). The improvement over NLM (76.03) is not large, but we believe this is a promising result nevertheless; it implies that our joint objective does indeed complement the strong results of distributional semantics. 
\end{enumerate}

\section{Conclusions and Future Work}

We advocate for a word representation learning algorithm that jointly implements both distributional and relational semantics. 
Our first contribution is an investigation of the ADMM algorithm, which flexibly combines multiple objectives. We show that ADMM converges quickly and is an effective method for combining multiple sources of information and linguistic intuitions into word representations. Note that other approaches for combining objectives besides ADMM are possible, including direct gradient descent on the joint objective, or concatenation of word representations individually optimized on independent objectives. A comparison of various approaches for multi-objective optimization in learning word representations, where the optimization space is riddled with local optima, is worthwhile as future work.  

The second contribution is a preliminary evaluation of three specific instantiations of ADMM, combining the NLM distributional objective with Graph Distance, TransE, or NTN relational objectives. In both the knowledge base completion and dependency parsing tasks, we demonstrate that the combined objective provides promising minor improvements compared to the single objective case. In the analogy task, we show that the combined objective is comparable to the single objective, and learns a very different kind of word representation. 

To the best of our knowledge, some recent work \cite{xu14knowledge,yu14embedding,faruqui14retrofit} explored similar motivations as ours. The main differences are in their optimization methods (i.e.~gradient descent directly on the joint objective is used in \cite{xu14knowledge,yu14embedding}, while \citet{faruqui14retrofit} introduces a post-processing graph-based method) as well as alternate relational objectives, e.g.~\citet{yu14embedding} formulate a skip-gram objective where word embeddings are trained to predict other words they share relations with. A detailed comparison on the same datasets would be beneficial to understand the impact of these design choices. 

Compared to the large body of existing work in word representations (e.g.\ LSA, \cite{deerwester90lsa}, ESA  \cite{gabrilovich07}, SDS \cite{mitchell2008}), the promise of recent learning-based approaches is that they enable a flexible definition of optimization objectives. 
As future work, we hope to further explore objectives beyond distributional and relational semantics and understand what objectives work best for each target task or application.

\section*{Acknowledgments}
This work is supported by a Microsoft Research CORE Grant and JSPS KAKENHI Grant Number 26730121.
D.F. was supported by the Flinn Scholarship during the course of this work. We thank Haixun Wang, Jun'ichi Tsujii, Tim Baldwin, Yuji Matsumoto, and several anonymous reviewers for helpful discussions at various stages of the project. 

\pagebreak
\bibliography{mybib}
\bibliographystyle{iclr2015}

\end{document}